\def\year{2022}\relax
\documentclass[letterpaper]{article} 
\usepackage{aaai22}  
\usepackage{times}  
\usepackage{helvet}  
\usepackage{courier}  
\usepackage[hyphens]{url}  
\usepackage{graphicx} 
\usepackage{natbib}  
\usepackage{caption} 
\DeclareCaptionStyle{ruled}{labelfont=normalfont,labelsep=colon,strut=off} 
\usepackage{algorithm}
\usepackage{algorithmic}
\usepackage{times}
\usepackage{latexsym}
\usepackage{amsmath,amssymb,amsfonts}
\usepackage{mathtools}
\usepackage{breqn}
\usepackage{bm}    
\usepackage{subcaption}
\usepackage{multirow}
\usepackage{booktabs}
\usepackage{xcolor}
\usepackage[switch]{lineno} 

\usepackage{newfloat}
\usepackage{listings}
\floatstyle{ruled}
\newfloat{listing}{tb}{lst}{}
\floatname{listing}{Listing}

\title{Non-Autoregressive Translation with Layer-Wise Prediction and Deep Supervision}
\author{
  Chenyang Huang\thanks{Work partially done during an internship at ByteDance AI Lab.}$^{1,2}$, 
  Hao Zhou$^2$,
  Osmar R. Za\"iane$^{1}$,
  Lili Mou$^{1}$, 
  Lei Li$^{3}$
}

\affiliations{
  $^1$Department of Computing Science,\\ Alberta Machine Intelligence Institute (Amii),  University of Alberta\\
  $^2$ByteDance AI Lab \\
  $^3$University of California, Santa Barbara \\
  \tt{\{chenyangh,zaiane\}@ualberta.ca} \\ \tt{zhouhao.nlp@bytedance.com}  \\
  {\tt doublepower.mou@gmail.com} \\
  {\tt lilei@cs.ucsb.edu}

}

\begin{document}
\nocopyright 
\maketitle
\newcommand{\model}{DSLP{}}

\begin{abstract}
    How do we perform efficient inference while retaining high translation quality? Existing neural machine translation models, such as Transformer, achieve high performance, but they decode words one by one, which is inefficient. Recent non-autoregressive translation models speed up the inference, but their quality is still inferior. In this work, we propose DSLP, a highly efficient and high-performance model for machine translation. The key insight is to train a non-autoregressive Transformer with \textbf{D}eep \textbf{S}upervision and feed additional \textbf{L}ayer-wise \textbf{P}redictions. We conducted extensive experiments on four translation tasks (both directions of WMT'14 EN--DE and WMT'16 EN--RO). Results show that our approach consistently improves the BLEU scores compared with respective base models. Specifically, our best variant outperforms the autoregressive model on three translation tasks, while being 14.8 times more efficient in inference.\footnote{Our code, training/evaluation scripts, and output are available at \url{https://github.com/chenyangh/DSLP}. \label{foot:link}} 
\end{abstract}

\section{Introduction}
Most state-of-the-art neural machine translation (NMT) systems are autoregressive, where they predict one word after another to generate the target sequence \cite{BahdanauCB14,attentionisallyouneed}. However, autoregressive NMT is slow in inference, which sometimes does not meet the efficiency requirement from the industry. 

\citet{gu2018nonautoregressive} propose non-autoregressive neural machine translation (NAT), which is 15.6 times faster than its autoregressive counterpart. While NAT achieves fast inference by generating target tokens in parallel, it assumes that the generated words are conditionally independent given the input. Such an independence assumption, however, weakens the power of sequence modeling, and results in worse performance than autoregressive translation. 

In recent studies, researchers propose partially \mbox{(non-)autoregressive} models by progressively generating several words at a time~\cite{ghazvininejad-etal-2019-mask} or iteratively editing with the Levenshtein Transformer~\cite{lev-transformer}. However, their inference is significantly slower than a full NAT model.
As shown by \citet{kasai2021deep}, 
a comparable-sized autoregressive model with a shallow decoder is able to outperform these approaches with similar latency. Therefore, these approaches do not achieve a desired quality--efficiency trade-off.

\begin{figure}[t]
 \centering
  \includegraphics[width=.9\linewidth]{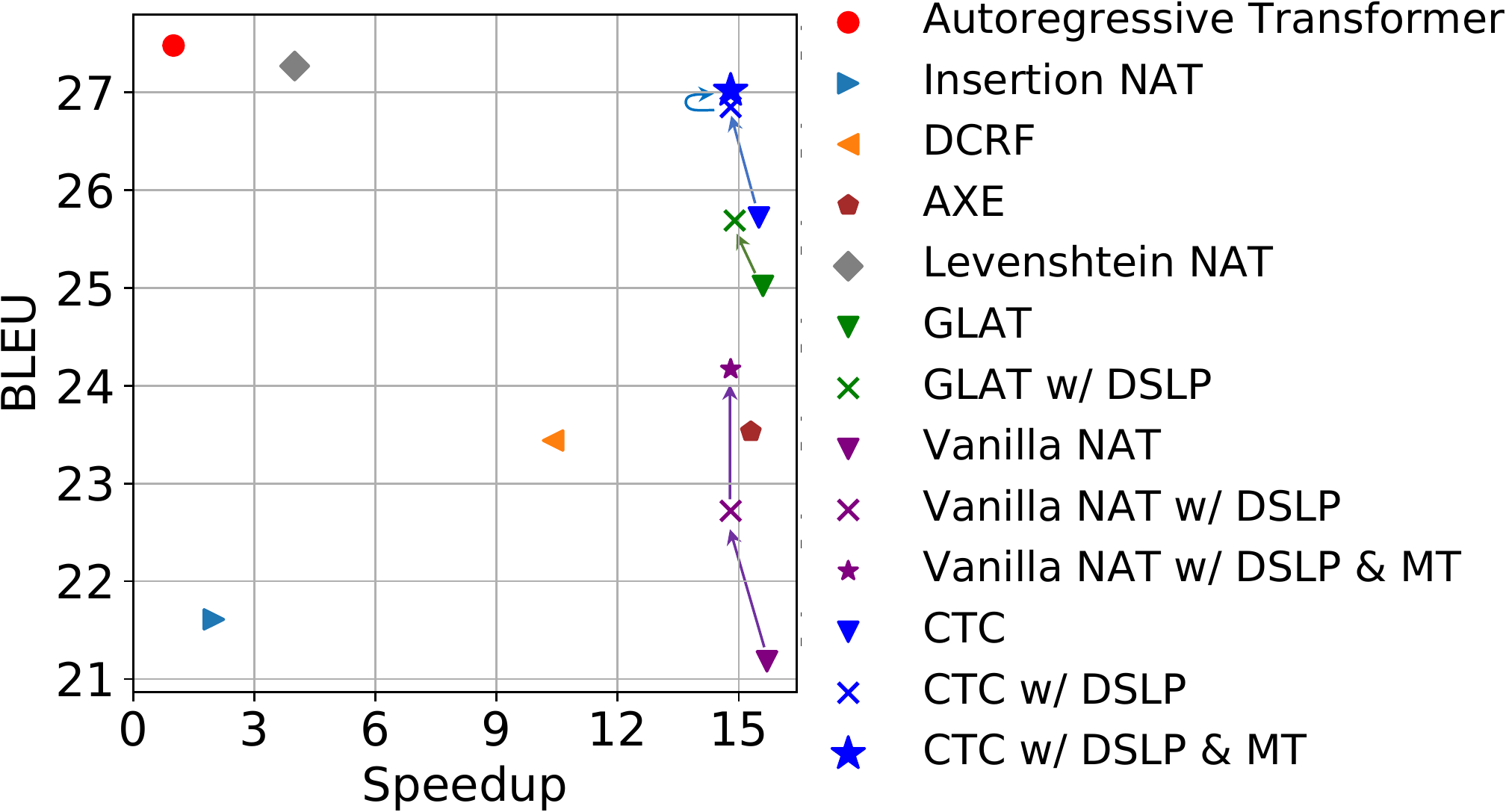}
  \centering
  \vspace{-.2cm} 
  \caption{Quality--efficiency trade-off for state-of-the-art NAT models and our \model{} on the WMT'14 EN--DE dataset. A cross ``$\times$'' represents our \model\ variants, and a star ``$\star$'' refers a DSLP model that is enhanced with our mixed training (MT) technique. Its base model is shown in the same color, and its correspondence is represented by an arrow.}
  \label{fig:all_trade_off}
  \vspace{-0.5cm}
\end{figure}

In this work, we propose a simple yet effective approach to non-autoregressive generation with a Deeply Supervised, Layer-wise Prediction-aware (\model) Transformer.
It is noticed that a traditional NAT Transformer~\cite{attentionisallyouneed} only makes predictions at the last layer, where all the words are generated in parallel. 
Thus, the prediction of a word is unaware of other time steps, which typically leads to inconsistent sentences. For example, the English phrase \textit{thank you} has two translations in German: \textit{danke sch{\"o}n} and \textit{vielen dank}. Instead of generating either of the phrases, a non-autoregressive model may predict \textit{danke dank}, which is absurd. 
Therefore, we propose a layer-wise prediction-aware Transformer that predicts the output at every decoder layer. The prediction is fed to the next Transformer layer for further processing. Hence, our decoder model is aware of different steps' predictions and can calibrate the NAT output through its decoder layers.
We further introduce deep supervision for training our prediction-aware decoder. This is essential to our model, because otherwise the intermediate predictions are not grounded to the target output and the calibration would be less meaningful. During training, we also propose to mix the intermediate predictions with groundtruth tokens, as the layer-wise predictions may still be meaningless after deep supervision. 

Our \model\ is a generic framework that can be combined with different base NAT models. In our experiments, we  evaluated \model{} on WMT'14 English--German and WMT'16 English--Romanian pairs, and considered the translation of both directions. Results show that our \model{} consistently improves the NAT performance for every translation language pair and every base model, 
including vanilla NAT, CMLM, GLAT, and CTC (see experiments section for details), demonstrating the generality of our approach. In addition, we show that mixing the layer-wise predictions with groundtruth during training further improves the performance. 
Remarkably, our best model, CTC with DSLP and mixed training, achieves better performance compared with its autoregressive teacher model on three of the four datasets with a 14.8x speedup. Figure~\ref{fig:all_trade_off} positions \model\ in prior work in terms of quality--efficiency trade-off.
\section{Related Work}

Non-autoregressive neural machine translation is attracting an increasing attention in the community~\cite{SunLWHLD19,lev-transformer,saharia-etal-2020-non,qian2020glancing}, because it is more efficient in inference than an autoregressive model, and has high potentials in industrial applications. Compared with autoregressive generation, NAT faces two major challenges: 1) determining the output length, and 2) handling multi-modality, i.e., generating a coherent text when multiple translations are plausible.

\citet{gu2018nonautoregressive} propose a token-fertility approach that predicts the repetition of source tokens to obtain the target length.
\citet{lee-etal-2018-deterministic} predict the length difference between the source and target sequences.

The Insertion Transformer \cite{stern} dynamically expands the canvas size (generation slots) through multiple rounds of inference. The Levenshtein Transformer \cite{lev-transformer} generates a sentence by explicitly predicting edit actions, such as insertion and deletion.
Alternatively, 
an empty token can be introduced into the vocabulary for generating a sentence of appropriate length \cite{ctc,saharia-etal-2020-non}.

To address the second challenge, \citet{gu2018nonautoregressive} propose a knowledge distillation (KD) approach that first trains an autoregressive translation (AT) model, and then trains the NAT model by AT's output instead of the groundtruth. 
\citet{Zhou2020Understanding} show that AT's output has fewer modes than the original corpus, and thus is easier for training NAT models.

Partially (non-)autoregressive generation is also widely used to mitigate the multi-modality issue, such as insertion-based methods~\cite{stern}, and the Levenshtein Transformer~\cite{lev-transformer}. 
\citet{ghazvininejad-etal-2019-mask} propose to pretrain a conditional masked language model (CMLM) for NAT, where they progressively generate several new words conditioned on previous ones.
\citet{wang-etal-2018-semi-autoregressive} model the dependencies among the words in a chunk. 
\citet{SunLWHLD19} propose to apply the linear-chain conditional random field on top of NAT predictions to capture token dependency. 
Their method is also partially autoregressive as beam search is required during inference. 

Recently, \citet{qian2020glancing} propose the Glancing Transformer (GLAT), which trains NAT predictions in a curriculum learning. Specifically, their method is built upon CMLM, and the amount of masked out groundtruth is dependent on the model's current performance. Thus, the model learns easy tokens first, and gradually moves on to hard ones. This also alleviates the multi-modal problem.

A similar study to ours is \citet{lee-etal-2018-deterministic}. They train two decoders: one is for the initial generation, and the other is applied iteratively for refinement. They also propose a deterministic variant of token fertility. 

Different from previous work, we perform layer-wise prediction within a decoder. This is an insightful contribution, being the key to efficient inference, as previous evidence shows that decoder-level iterative refinement does not have an advantage over a carefully designed autoregressive model~\cite{kasai2021deep}.
We also show that our layer-wise prediction awareness is agnostic to base NAT models, being a generic framework.

Our work requires deep supervision on the Transformer decoder. This is similar to the treatment in the depth-adaptive Transformer \cite[DAT,][]{depthadaptive}, where they apply deep supervision for an autoregressive model to allow ``early exit,'' i.e., using a subset of decoder layers. Our NAT work is apparently different from DAT.

\textbf{Discussion on autoregressiveness.} At first glance, it is doubtful whether our \model\ is non-autoregressive, as we have layer-wise prediction. In statistics, \textit{autoregression} means using past predictions for future ones, typically in a time series~\cite{AR}. Our \model\ differs from partially autoregressive models~\cite{stern,lev-transformer,ghazvininejad-etal-2019-mask}. We predict the output words simultaneously in the last layer, whereas its layer-wise prediction is an internal mechanism, similar to Transformer hidden states. Thus, \model\ should be considered as a full NAT model from the perspective of statistics.

More importantly, the motivation of NAT research is the inference efficiency for industrial applications. Being non-autoregressiveness itself is not the research goal. As demonstrated in experiments, our work achieves satisfactory quality--efficiency trade-off, as it does not cause much extra computational cost.

\section{Methodology}

In this section, we describe our model in detail. We first introduce a generic Transformer-based NAT model. Then, we present our layer-wise prediction-aware approach and the deeply supervised training. We also present an improved training scheme that takes groundtruth tokens as input and mixes them the layer-wise predictions. 

\begin{figure}[t]
  \includegraphics[width=0.8\linewidth]{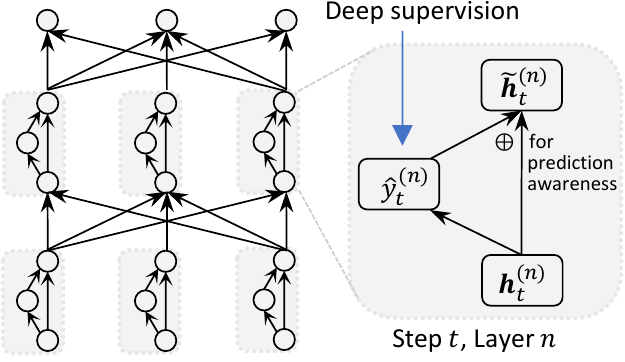}
  \centering
  \caption{Overview of our Deeply Supervised Layer-wise Prediction-aware (\model) Transformer.}
  \label{fig:motivation}
  \vspace{-0.3cm}
\end{figure}

\subsection{Non-Autoregressive Transformer}\label{ss:nat}

Recent advances in neural machine translation are built upon the Transformer architecture \cite{attentionisallyouneed}. It is an encoder--decoder neural network, where the encoder is responsible for representing the source sentences, and the decoder is used to generate the target translation.

Formally, let the source text be $\mathbf x = (\mathtt x_1, \cdots, \mathtt x_{T_{x}})$, where $T_x$ is the length.
The encoder first takes $\bf x$ as input and maps these discrete words\footnote{We adopt BPE segmentation, and strictly speaking, they should be tokens. For clarity, we use \textit{word} and \textit{token} interchangeably in the paper.} into vector representations by an embedding function, denoted by $\bm e^{(0)}_t = \operatorname{emb}(\mathtt x_t)$. Each Transformer layer performs multi-head attentions with a feed-forward layer to compute deep contextual representation, denoted by
\begin{align}
    \bm e_{1:T_x}^{(n)} & = \operatorname{Layer}^{(n)}_{\text{enc}}(\bm e_{1:{T_{x}}}^{(n-1)}) 
\end{align}
where $\operatorname{Layer}_{\text{enc}}^{(n)}$ is the $n$th encoder layer.
Such processing is repeated for every layer, and we take the last layer as the encoder's hidden representation, denoted by $\mathbf E$.

For non-autoregressive decoding, the first step is to determine the length $T_y$ of the output sentence. This can be handled in various ways. For example, \citet{gu2018nonautoregressive} train a classifier to predict the length during inference, and in training, the length is given by the groundtruth. In latent alignment models~\cite{saharia-etal-2020-non}, an empty prediction is allowed, and $T_y$ is typically set as $kT_x$ for a constant $k$. Our \model{} approach can be built upon multiple base models, and the output length is handled accordingly.

For the decoder input $\bm h_{t}^{(0)}$, we follow \citet{saharia-etal-2020-non} and feed the embedding of a special token $\rm s$ for every step. In our preliminary experiments, this treatment is on par with copying encoder representations as the decoder input~\cite{gu2018nonautoregressive,wei-etal-2019-imitation}.

The decoder processes the information in a similar way to the encoder, except that it has an additional encoder--decoder attention to obtain input information. Thus, a decoder layer can be represented by
\begin{align}
    \bm h_{1:T_y}^{(n)} & = \operatorname{Layer}^{(n)}_{\text{dec}}(\bm h_{1:T_y}^{(n-1)}, \bf E) \label{eq:nat_h} 
\end{align}
where $\operatorname{Layer}^{(n)}_{\text{dec}}$ is the $n$th decoder layer.

Finally, NAT uses a softmax layer to predict the target words based on the decoder's last hidden states, given by
\begin{align}
    p(\mathrm y_t | \bm h_t^{(N)}) &= \operatorname{softmax} (\mathbf W \bm h_t^{(N)}) \label{eq:nat_pred}
\end{align}
where $N$ is the number of decoder layers. In this way, the output words $\mathtt y_1, \cdots, \mathtt y_{T_y}$ are predicted simultaneously in a non-autoregressive fashion.

The training objective is to maximize the log-likelihood 
\begin{align}
	\textstyle
    \mathcal{L} = \sum_{t=1}^{T_y} \log p(\mathrm y_t | \bm h_t^{(N)}) \label{eq:nat_loss}
\end{align}

\subsection{Layer-Wise Prediction-Awareness}\label{ss:PA}
As we observe from~(\ref{eq:nat_pred}), when a conventional NAT model predicts a word $\mathrm y_t$, without being aware of the other words $\mathrm y_1, \cdots,  \mathrm y_{t-1},\mathrm y_{t+1},\cdots, \mathrm y_{T_y}$. This is undesired when the target distribution is ``multi-modal,''  i.e., multiple outputs are appropriate given an input.

To this end, we propose a layer-wise prediction-aware decoder that generates an output sequence at every layer. Such an output serves as a tentative translation, and is fed to the next layer for further processing. 

Consider the $t$th step of the $n$th decoder layer. 
We perform linear transformation on the conventional Transformer hidden state  ${\bm{h}}^{(n)}_t$, and use softmax to predict the probability of words:
\begin{align}
    p(\mathrm y^{(n)}_t | {\bm{h}}^{(n)}_t) &= \operatorname{softmax} (\mathbf W {\bm{h}}_t^{(n)})\label{eq:prob}
\end{align}

Then, we obtain the most probable word by 
\begin{align}
    \widehat{\mathrm y}_t^{(n)} & = \operatorname{argmax} p(\mathrm y_t^{(n)} | {\bm{h}}^{(n)}_t) \label{eq:lp_argmax} 
\end{align}
and concatenate its embedding with $\bm h^{(n)}_t$ and further processed by a linear layer
\begin{align}
     \widetilde{\bm{h}}_t^{(n)}  & = \mathbf W_c [{\bm{h}}_t^{(n)}; \operatorname{emb}(\widehat{\mathrm y}_t^{(n)} )] \label{eq:lp_concat}
\end{align}
where $\mathbf W_c$ is a weight matrix. $\widetilde{\bm{h}}_t^{(n)} $ is an updated prediction-aware hidden state, fed to the next layer. Notice that the last layer does not require such an update, and the final output is directly given by (\ref{eq:prob}).

In this way, the prediction at the $(n+1)$th layer is aware of all the predictions at the $n$th layer, due to the attention mechanism in the Transformer architecture. This serves as a calibration mechanism that can revise the tentative generation through multiple decoder layers.

\subsection{Deeply Supervised Training}
\label{ss:DS}
We further propose to train our layer-wise prediction-aware Transformer with deep supervision.

Typically, NAT models are trained with the supervision at the last layer. With such training, however, our intermediate predictions are not grounded to the desired output sentence, and such prediction-awareness becomes less meaningful.

Specifically, we apply maximum likelihood estimation to the prediction of every decoder layer, given by
 \begin{align}
 \textstyle
    \mathcal{L}_D = \sum_{n=1}^N \sum_{t=1}^{T_y} \log p(\mathrm y^{(n)}_t | \widetilde{\bm{h}}^{(n)}_t) \label{eq:per_layer_nat_loss}
\end{align}

This layer-wise deep supervision ensures that our \model{} predicts the target output (although imperfect) at every decoder layer. By feeding back such prediction, the model is able to calibration the words of different time steps, alleviating the weakness of NAT models. Figure~\ref{fig:motivation} presents an overview of our DSLP.

\subsection{Mixing Predictions and Groundtruth in Training}
\label{sec:ss}
During training, the layer-wise predictions can still be of low quality despite deep supervision, in which case the next layers may not be well trained. 
As a remedy, we propose to partially feed the groundtruth to intermediate layers.

Formally, as opposed to (\ref{eq:lp_concat}), we compute the hidden states of next layers as,
\begin{align}
     \widetilde{\bm{h}}_t^{(n)}  & = \mathbf W_c [{\bm{h}}_t^{(n)}; \operatorname{emb}(\overline{\mathrm y}_t^{(n)} )] \label{eq:lp_ss}
\end{align}
where $\overline{\mathrm y}_t^{(n)} = s_t {\mathrm y}_t + (1-s_t) \widehat{\mathrm y}_t^{(n)}$, and $s_t \sim \operatorname{Bernoulli}(\lambda)$. $\lambda$ is a hyperparameter controlling the amount of mixed groundtruth tokens, referred to as the mixing ratio.
Note that $s_t$ is not dependent on the layer number $n$, otherwise we may feed the model with all the groundtruth tokens during training and having low performance.

Although our \model{} is generic to non-autoregressive models, the proposed mix training scheme needs to be adapted to each NAT model accordingly. 

For the vanilla NAT \cite{gu2018nonautoregressive}, we mask out the observed groundtruth tokens in the loss, which prevents the model from learning simply copying tokens. This is similar to the masked language models \cite{devlin-etal-2019-bert,ghazvininejad-etal-2019-mask}\footnote{Empirically, we find the masked improves the BLEU score by 0.2 on WMT 14' EN$\rightarrow$DE. However, the masked loss is not our main focus, and our mixed training improves upon the the vanilla NAT-based DSLP model by 1.4 BLEU (see Table~\ref{tab:dslablation_results}).}.

For the connectionist temporal classification \cite[CTC,][]{ctc} model, the predictions are usually longer than the groundtruth. This is because CTC allows predicting extra tokens (including repeated tokens and empty tokens) to generate sentences of different lengths, and the probability of generating the groundtruth sequence $\mathrm y_{1:T_y}$  is computed by marginalizing all valid alignments\footnote{An alignment is an expanded sequence that can be reduced to the groundtruth sequence. For example, ``\_\_aabb\_c'' is a valid alignment to the sequence ``abc'', where ``\_'' represents an empty token. The marginalization of all valid alignments is computed efficiently via dynamic programming. More details can be found in Section 3 of \citet{ctc}.}. The length discrepancy prevents us from directly replacing a predicted token with a groundtruth token. Alternatively, we regard the best alignment given by the model as the pseudo groundtruth. This is similar to how \citet{gu-kong-2021-fully} combine CTC with the Glancing Transformer \cite[GLAT,][]{qian2020glancing}.

The mixing training does not improve the models that have already used groundtruth during training, such as the conditional masked language model \cite[CMLM,][]{ghazvininejad-etal-2019-mask} and GLAT. This is because further feeding groundtruth tokens interferes with their training schemes. 

\vspace{-0.1cm}
\subsection{Inference}

The inference of our \model{} resembles its training process: the predicted output of every layer is fed to the next layer, and we take the last layer as the final output. 
It should be pointed out that our layer-wise prediction does not introduce much extra computational cost. The  experiments section will show that, despite the high translation quality of \model{}, the time difference between \model{} and a vanilla NAT model is negligible.

\begin{table*}[t]
\vspace{-.1cm}
\centering
\resizebox{.7\linewidth}{!}{
\begin{tabular}{rl|cc|cc|c|c} \toprule
           \multirow{2}{*}{Row\#}   &\multirow{2}{*}{Model}& \multicolumn{2}{c|}{WMT'14} & \multicolumn{2}{c|}{WMT'16} & \multirow{2}{*}{Latency (ms)} & \multirow{2}{*}{Speedup} \\
             &    & EN--DE        & DE--EN       & EN--RO           & RO--EN    &                          &                           \\ \midrule
 1  & Transformer (teacher)         & 27.48        & 31.21       & 33.70           & 34.05    &     326.80                 &        1$\times$                     \\ \midrule
2 & CMLM$_1$     & 19.91        &  22.69      &   27.60         & 29.00    &   20.91   &    15.6$\times$     \\ 
3 & ~~ w/ \model{}  & \textbf{21.76}        &     \textbf{25.30}  &   \textbf{30.29}         & \textbf{30.89}    &      21.76    &   15.0$\times$                         \\ \midrule
4 & GLAT       & 25.02        & 29.63       &   31.33         & 32.43    &   20.91            &       15.6$\times$     \\
5 & ~~ w/ \model{}    & \textbf{25.69}        & \textbf{29.90}       &   \textbf{32.36}         & \textbf{33.06}    &   21.73        &  14.9$\times$                         \\ \midrule
6 & Vanilla NAT        & 21.18        & 24.93       &   29.15         & 29.69    &     20.86                 &         15.7$\times$                   \\
7 & ~~ w/ \model{}  & 22.72        & 25.83       &  30.48     & 31.46    &  22.12           &    14.8$\times$    \\
8 & ~~ w/ \model{} \& Mixed Training & \textbf{24.17}        & \textbf{28.63}       &  \textbf{31.49}     & \textbf{32.64}    &  22.12           &    14.8$\times$                     \\ \midrule
9 & CTC        & 25.72        & 29.89       &   32.89         & 33.79    &       21.04            &       15.5$\times$         \\
10 & ~~ w/ \model{}         & 26.85        & 31.16       &   33.85         & 34.24    &       22.06            &       14.8$\times$        \\
11 & ~~ w/ \model{} \& Mixed Training & \textbf{27.02}        & \textbf{31.61}       &   \textbf{34.17}         & \textbf{34.60}   &   22.06       &   14.8$\times$                         \\  \midrule
& Average Improvement   &  1.34    &  1.38        & 1.58    &  1.28   &   -- & --       \\ 
\bottomrule
\end{tabular}
}
\vspace{-.1cm}
\caption{Applying \model\ to different base NAT models, which shows the generality of our approach. Translation quality is evaluated in BLEU. Latency is the processing time (in milliseconds) of a single sentence. Speedup is relative to an autoregressive model.  All results are based on our implementation. CMLM$_k$ refers to $k$ iterations of progressive generation. Here, we consider $k=1$, as more iterations make CMLM closer to autoregressive models.}
\label{tab:dslablation_results}
\vspace{-.1cm}
\end{table*}

\begin{table*}[t]
\centering
\resizebox{.85\linewidth}{!}{
\begin{tabular}{l|rl|cc|cc|c} \toprule
             \multirow{2}{*}{Category} &\multirow{2}{*}{Row\#} & \multirow{2}{*}{Model}& \multicolumn{2}{c|}{WMT'14} & \multicolumn{2}{c|}{WMT'16} &  \multirow{2}{*}{Speedup} \\
            & &  & EN--DE        & DE--EN       & EN--RO           & RO--EN    &                                        \\ \midrule
Autoregressive &1 &   Transformer          & 27.48        & 31.21       & 33.70           & 34.05   &        1$\times$                     \\ \midrule
Iterative &2 & Iterative Refinement \cite{lee-etal-2018-deterministic} & 21.61      & 25.48 & 29.32  & 30.19     &  \ \ 2.0$\times$  \\
& 3 &  Blockwise \cite{blockwise}   &   27.40   &   --  &   --  & --   & \ \  3.0$\times$  \\ 
& 4 &  Insertion Transformer \cite{stern} &  27.41   &  -- & --  &  --   & \ \ 4.8$\times$  \\
& 5 &  Levenshtein Transformer \cite{lev-transformer} &  27.27  &  --       &  --   & --   & \ \ 4.0$\times$  \\  
& 6 &  CMLM$_{10}$ \cite{ghazvininejad-etal-2019-mask}  & 27.03      & 30.53       & 33.08           & 33.08        & \ \ \ 2.6$\times$$^\ddagger$     \\
& 7 &  Imputer \cite{saharia-etal-2020-non} & \textbf{28.2}  & \textbf{31.8} &   \textbf{34.4}  &  34.1     & \ \ 3.9$\times$  \\
& 8 &  DisCO \cite{kasai2021deep}   & 27.34 &  31.31     &  33.22 &     33.25  & \ \ \ 3.5$\times$$^\dagger$  \\ \midrule
Non-iterative& 9 & Vanilla NAT  \cite{gu2018nonautoregressive}    & 17.69       & 21.47       & 27.29           & 29.06    &       15.6$\times$                   \\
& 10 & DCRF \cite{SunLWHLD19}    &   23.44     &   27.22 &   --      &       --   &  10.4$\times$  \\
& 11 & CMLM$_1$ \cite{ghazvininejad-etal-2019-mask}  & 18.05       & 21.83       & 27.32           & 28.20          & \ 15.6$\times$$^\ddagger$   \\
& 12 & AXE  \cite{GhazvininejadKZ20} &  23.53  &     27.90 & 30.75   & 31.54    & 15.3$\times$  \\ 
& 13 & CTC \cite{saharia-etal-2020-non}       & 25.7        & 28.1       & 32.2           & 31.6     &       18.6$\times$             \\
& 14 & CNAT \cite{bao-etal-2021-non} & 25.67       & 29.36       & --           & --             &     10.4$\times$       \\
& 15 & GLAT  \cite{qian2020glancing}& 25.21        & 29.84       & 31.19           & 32.04             &     15.3$\times$         \\ 
& 16 & CTC + GLAT  \cite{qian2020glancing}& 26.39        & 29.54       & 32.79           & 33.84             &     14.6$\times$         \\ 
& 17 & CTC + VAE \cite{gu-kong-2021-fully} & 27.49     & 31.10       & 33.79           & 33.87             &     16.5$\times$         \\
& 18 & CTC + GLAT \cite{gu-kong-2021-fully} & 27.20        & 31.39       & 33.71           & 34.16             &     16.8$\times$       \\ \midrule
Ours & 19 & CTC w/ DSLP \& Mixed Training & 27.02       & 31.61       &   34.17         & \textbf{34.60}   &   14.8$\times$                        \\  
\bottomrule
\end{tabular}
}
\vspace{-0.1cm}
\caption{Comparing our model with state-of-the-art NAT models. Results of prior work are quoted from respective papers. $^\dagger$~indicates that the number is estimated from the plot in the previous paper. $^\ddagger$~indicates that the inference time in \citet{ghazvininejad-etal-2019-mask} is not available. Their models' efficiency is given by our own implementation.}
\label{tab:main_results}
\vspace{-.2cm}
\end{table*}

\section{Experiments}\label{sec:experiment}
\subsection{Datasets}
We evaluated our models on benchmark translation datasets: WMT'14 English--German (4.0M sentence pairs) and WMT'16 English--Romanian (610K pairs). For fair comparison, we obtain the preprocessed corpus (tokenization and vocabulary) released by previous work: \citet{Zhou2020Understanding} for WMT'14 EN--DE, and \citet{lee-etal-2018-deterministic} for WMT'16 EN--RO.
We consider both translation directions, and in total, we have 4 translation tasks.

\subsection{Experimental Setup}
\textbf{Hyperparameters.}
We mostly followed the standard hyperparameters used in NAT research. We used Transformer \cite{attentionisallyouneed}: both the encoder and the decoder had 6 layers, each layer had 8 attention heads, and the hidden dimension was 512 for attention modules and 2048 for feed-forward modules. 
To train the models, we used a batch size of 128K tokens, with a maximum 300K updates. 
For regularization, we set the dropout rate to 0.1 for EN--DE and 0.3 for EN--RO. We applied weight decay of 0.01 and label smoothing $0.1$. To obtain robust results, we averaged the last 5 best checkpoints, following \citet{attentionisallyouneed}. For the mixed training, we used a fixed mixing ratio $\lambda$ and set it to 0.3. More details can be found in our anonymized GitHub repository (Footnote \ref{foot:link}).

\textbf{Knowledge Distillation.} It is a common technique to train NAT by an autoregressive model's output. Previous evidence shows that this largely improves NAT performance~\cite{gu2018nonautoregressive,lee-etal-2018-deterministic,stern}. In our study, we also adopted knowledge distillation from an autoregressive Transformer model for all baselines and our \model{} variants. 

\textbf{Evaluation Metrics.}
For translation quality, we computed BLEU \cite{papineni-etal-2002-bleu} scores over tokenized sentences. 
To measure inference latency, we used a single Nvidia V100 GPU and perform inference with one sentence at a time. This mimics a deployed NMT system in the industry.
Our models were implemented and evaluated with the open-source toolkit \texttt{FairSeq} \cite{fairseq}.

\textbf{Base Models.} 
Our \model{} can be applied to various base NAT models. 
To evaluate its generality, we consider the following models: 1) Vanilla NAT \cite{gu2018nonautoregressive},  which is the foundation of NAT models. 2)  Partially non-autoregressive generation based on a conditional masked language model \cite[CMLM,][]{ghazvininejad-etal-2019-mask}, which progressively generates several words of a sentence. CMLM$_k$ refers to $k$ iterations. 3) Glancing Transformer \cite[GLAT,][]{qian2020glancing}, which progressively trains NAT word predictors in a curriculum learning fashion. 4) Connectionist temporal classification \cite[CTC,][]{ctc}, which allows empty tokens and performs dynamic programming for marginalization of latent alignment.
With these base models, we would be able to evaluate \model{} with a wide range of prediction schemas and training objectives.

\subsection{Main Results}
\textbf{Generality of \model.}
Table~\ref{tab:dslablation_results} shows the performance of our \model\ compared with base models: Vanilla NAT, CMLM, GLAT, and CTC. Comparing with results published in previous papers~(quoted in Table~\ref{tab:main_results}), our replications mostly match the previous work, except that our vanilla NAT outperforms the implementation of \citet{gu2018nonautoregressive}. It is probably due to the engineering efforts by the community for NAT in the past years. The results on base models implicate that our implementation is fair for our study on \model.

We apply \model{} to all baseline models. As seen, \model{} consistently achieves higher performance for every base model and every translation task. \model{} consistently outperforms the base model by more than 1 BLEU point, which is a considerable improvement over strong baselines for machine translation.

\textbf{Improving DSLP by Mixed Training.} 
For the training of vanilla NAT- and CTC-based DSLP, we mix the layer-wise predictions with groundtruth tokens. 
We do not apply the mixed training on CMLM and GLAT, as they have already used groundtruth tokens in training, and further feeding groundtruth causes potential conflicts. 

We show the improvement of the mixed training in Table~\ref{tab:dslablation_results} (Lines 8 and 11). We observe mixed training consistently improves the performance, with an average of 0.97 BLEU improvement compared with respective DSLP variants.

When combined with the weak vanilla NAT model, DSLP with mixed training outperforms the strong baseline model, the Glancing Transformer \cite{qian2020glancing}, on EN$\rightarrow$RO, and RO$\rightarrow$EN datasets. When combining with the strong CTC model, the mixed training scheme further improves the performance consistently, even though the CTC with DSLP model has already achieves superb results.

\begin{figure*}[!t]
\vspace{-0.1cm}
\centering
    \includegraphics[width=.25\linewidth]{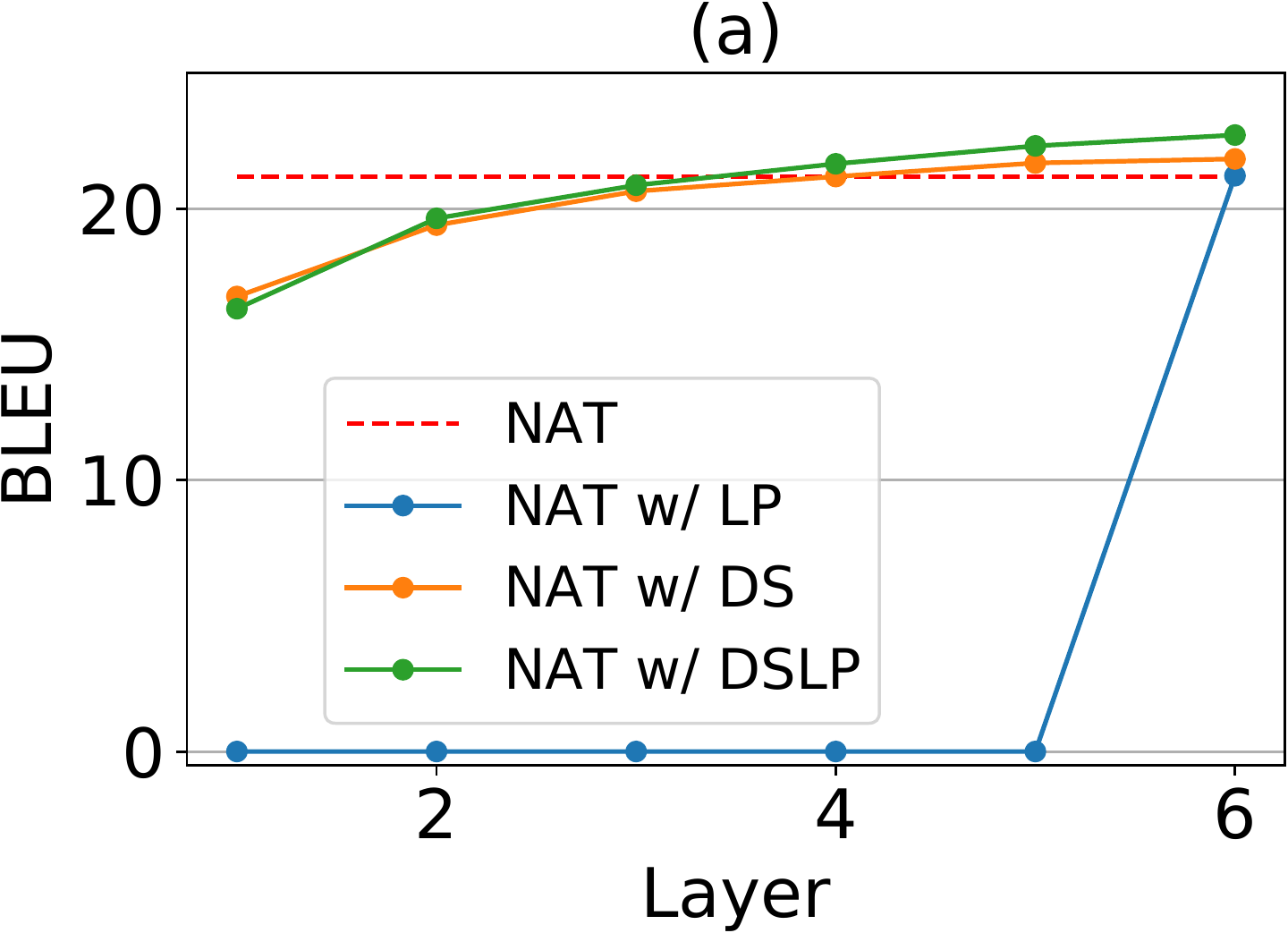}
    \includegraphics[width=.25\linewidth]{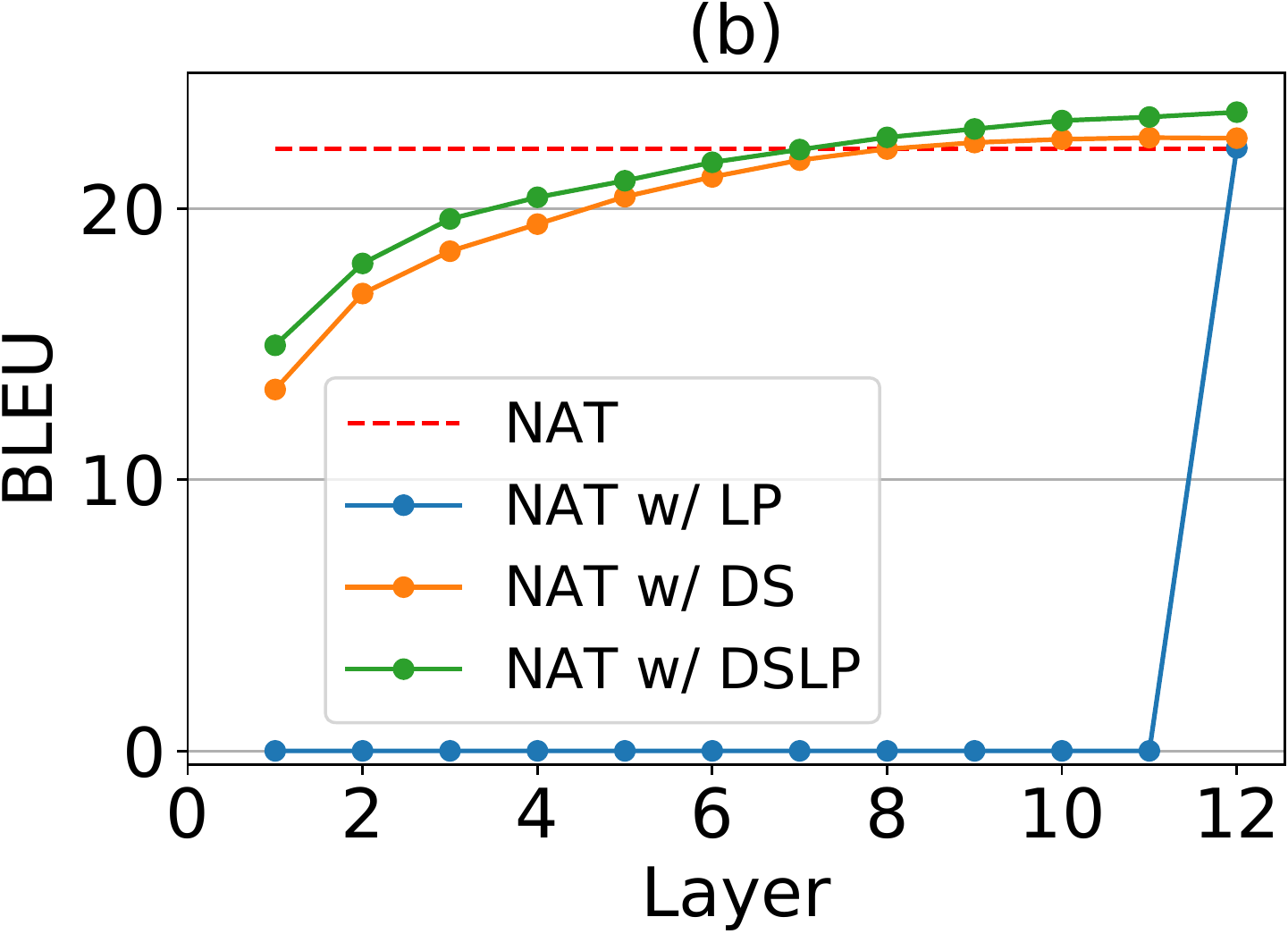}
    \includegraphics[width=.25\linewidth]{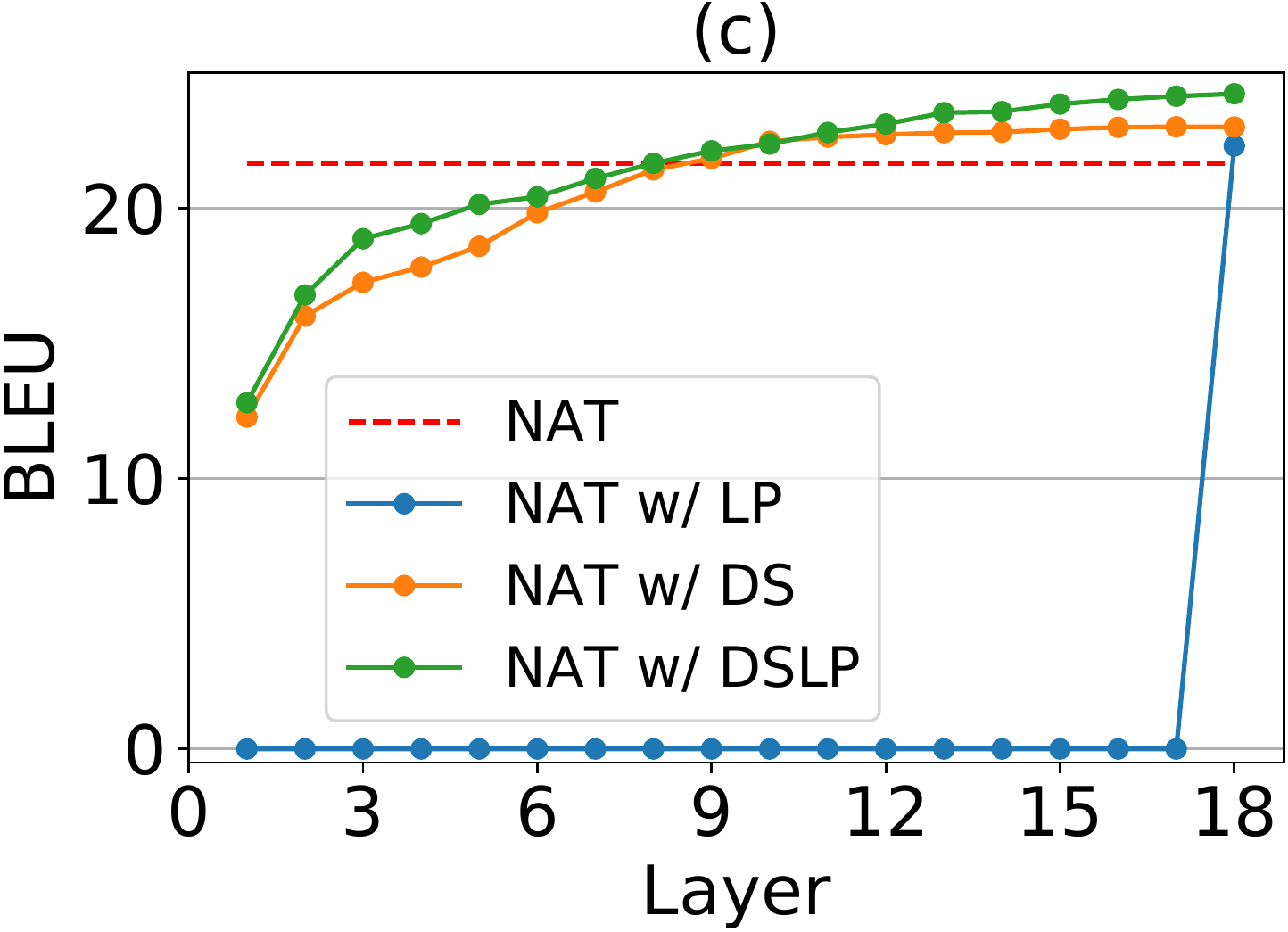}
\vspace{-.1cm}
\caption{Layer-wise performance of a fixed, trained vanilla NAT model. Decoder layers: 6 for (a), 12 for (b), and 18 for (c).}
\label{fig:per_layer}
\vspace{-0.1cm}
\end{figure*}

\begin{table}[!t]
\centering
\resizebox{.8\linewidth}{!}{
\begin{tabular}{llll} \toprule
           & 6      & 12        & 18        \\ \midrule
Vanilla NAT        & 21.18  &  22.22    &  21.63    \\
~~ \quad w/ LP   & 21.22{\scriptsize(+0.04)}  &  22.24{\scriptsize(+0.02)}    &  22.29{\scriptsize(+0.66)}    \\
~~ \quad w/ DS   & 21.84{\scriptsize(+0.66)}   &  22.66{\scriptsize(+0.38)}    &  22.99{\scriptsize(+1.36)}    \\
~~ \quad w/ DSLP & \textbf{22.72}{\scriptsize(+1.54)}   &  \textbf{23.56}{\scriptsize(+1.34)}     &  \textbf{24.22}{\scriptsize(+2.59)}    \\ \bottomrule    
\end{tabular}}
\caption{Ablation study on layer-wise prediction (LP) and deep supervision (DS). We tested models of 6 layers, 12 layers, and 18 layers. The number in bracket shows the improvement over vanilla NAT.}
\label{tab:ablation}
\end{table}

\textbf{Comparing with the State of the Art.} 
We compare our best variant (CTC w/ \model{} \& Mixed Training) with previous state-of-the-art NAT models in Table~\ref{tab:main_results}. Profoundly, our approach achieves close results to the autoregressive (AT) teacher model on the WMT 14' EN$\rightarrow$DE dataset, and even outperforms AT on other three translation datasets. 

Compared with iterative methods (Line 2--8, Table~\ref{tab:main_results}), our model produces very competitive translation quality while being approximately 4 times faster in inference. 
Compared with non-iterative methods (Line 9--16, Table~\ref{tab:main_results}), our CTC w/ \model{} \& Mixed Training outperforms all existing systems on three datasets (DE$\rightarrow$EN, EN$\rightarrow$RO, and RO$\rightarrow$EN), while only costing extra 4--6\% latency.

\subsection{Analysis}

In this part, we present in-depth analysis on \model{} and the mixed training. Our development was mainly conducted on vanilla NAT because we would like to rule out complications brought by various NAT models. It is also more efficient (and greener). Therefore, the base model is vanilla NAT for this subsection (unless otherwise stated).

\textbf{Ablation Study on DSLP.}
Table~\ref{tab:ablation} presents an ablation study on the layer-wise prediction (LP) and deep supervision (DS).
We observe that LP or DS alone shows little improvement over the vanilla NAT. This is understandable since LP alone does not guarantee that intermediate predictions are grounded to target, and such awareness is mostly ineffective. DS alone does not have a calibration mechanism of different time steps. It nevertheless improves the performance to a small extent when the model is deep. This is because deep Transformers are difficult to train~\cite{pmlr-v119-huang20f}, as the vanilla NAT's performance drops by 0.59 from 12 layers to 18 layers; DS may help training a deep structure.

Our \model{} significantly outperforms LP and DS on all the settings. Compared with DS, our full \model{} shows more improvement when the network is deep, as more layers allow larger calibration capacity of \model.

\begin{figure}[!t]
\vspace{-0.2cm}
\resizebox{\linewidth}{!}{
\mbox{\includegraphics[width=.5\linewidth]{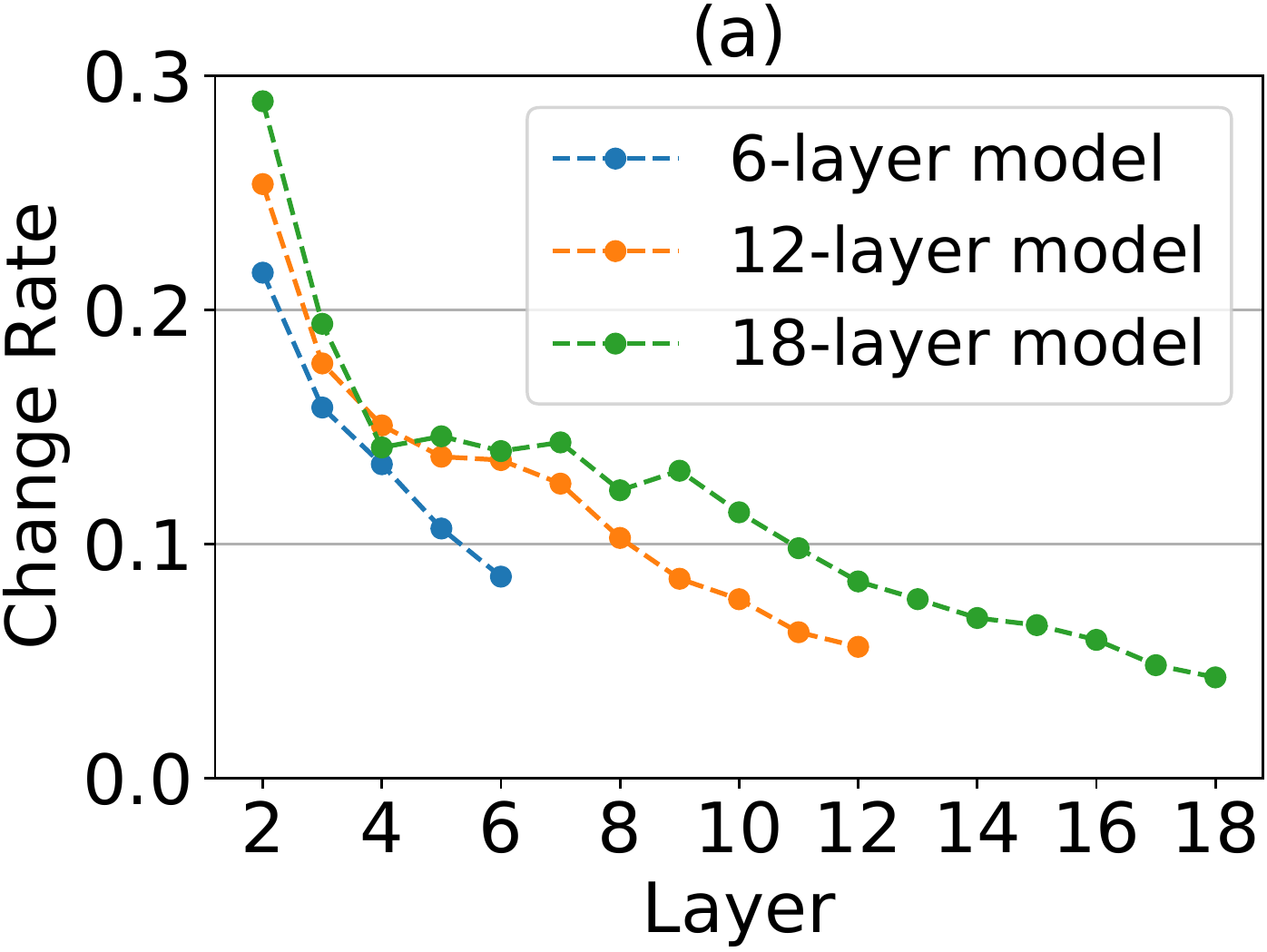}
\includegraphics[width=.5\linewidth]{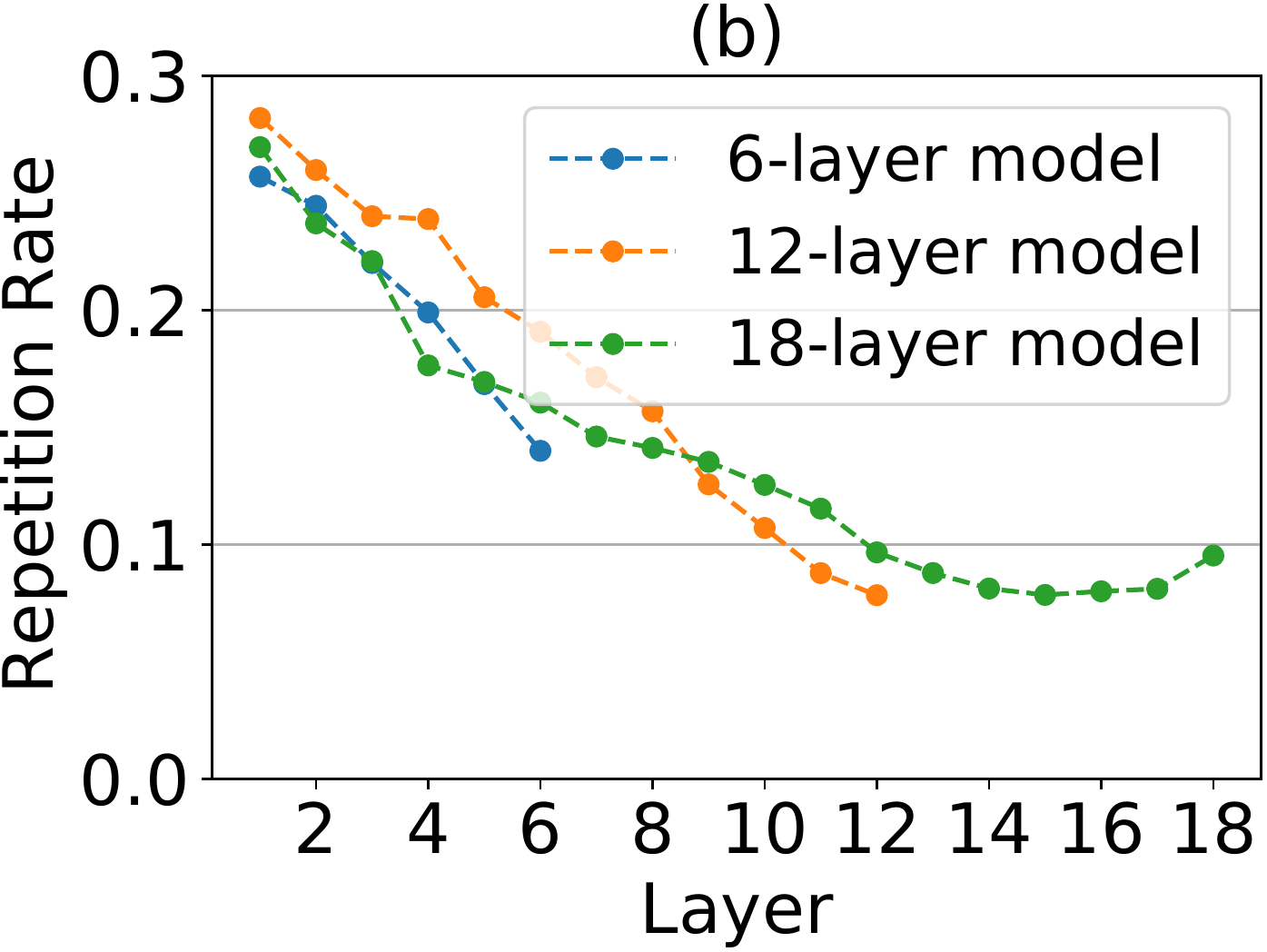}}}
   \vspace{-0.2cm}
 \caption{(a) Change rate for vanilla NAT w/ \model{}. The first layer is excluded as it does not perform editing. (b) Word repetition rate in each layer of vanilla NAT w/ \model.}
   \label{fig:layer}
\vspace{-0.2cm}
\end{figure}

\textbf{Layer-Wise Performance.}
Since our model makes intermediate predictions, we are curious how the performance evolves through the decoder layers.  We took a fixed, trained model, and tested the BLEU score of every layer. In Figure~\ref{fig:per_layer}, we show the performance dynamics of a 6-, 12-, and 18-decoder-layer models.

With only LP, the model achieves zero BLEU for all intermediate layers. 
This verifies that LP alone does not provide meaningful predictions at intermediate layers, leading to lower performance than our full \model{}. 

On the other hand, DS is able to achieve reasonable BLEU scores at every decoder layer. However, their BLEU scores are consistently lower than those of \model.

Notably, DS shows no improvement (or sometimes even decreases) at the last few layers. By contrast, the performance of \model{} improves steadily by the calibration over multiple layers.

\textbf{Change Rate.}
Since our \model{} performs layer-wise calibration, we investigate how much word changing is performed in every layer. Figure~\ref{fig:layer}a shows the percentage of changed words, where we also consider three variants: 6-layer, 12-layer, and 18-layer models.

We observe that, in each model, early layers perform more changes, and the change rate decreases gradually.
This is reasonable because the last few layers have produced high-quality translations,  requiring less calibration.

\begin{figure*}[!ht] \footnotesize
\centering
\vspace{-0.1cm}
\resizebox{0.95\linewidth}{!}{
\begin{tabular}{|l|l|}
\hline
    \rotatebox[origin=c]{90}{\textbf{\scriptsize Example 1}} & \resizebox{0.85\linewidth}{!}{
\begin{tabular}{rllllllllllllll}
\multicolumn{15}{l}{\textbf{Source: } Würde ich Gleichgesinnte erst nach Einbruch der Dunkelheit treffen können?  }\\
\multicolumn{15}{l}{\textbf{Reference: } Would I be forced to meet like-minded people only after dark? } \\

\textbf{Generation: \quad}Step: & 1 & 2 & 3 & 4 & 5 & 6 & 7 & 8 & 9 & 10 & 11 & 12 & 13 & 14\\
Layer 1: & Would & I & meet & meet & meet & minded & minded & minded & after & dark- & dark- & dark- & collapsed & ? \\
Layer 2: & Would & I & only & meet & like & minded & minded & minded & after & dark- & ness   & dark- & collapsed & ? \\
Layer 3: & Would & I & meet & meet & like & @-@    & minded & people & after & dark- & ness   & dark- & collapsed & ? \\
Layer 4: & Would & I & meet & meet & like & @-@    & minded & people & after & dark- & dark- & ness   & collapsed & ? \\
Layer 5: & Would & I & only & meet & like & @-@    & minded & people & after & dark- & ness   & ness   & collapsed & ? \\
Layer 6: & Would & I & only & meet & like & @-@    & minded & people & after & dark- & dark- & ness   & collapsed & ?
\end{tabular}} \\
\hline\hline
    \rotatebox[origin=c]{90}{\scriptsize\textbf{Example 2}}  & 
    \resizebox{0.85\linewidth}{!}{
\begin{tabular}{rlllllllllllllllll}
\multicolumn{18}{l}{\textbf{Source: } Konflikte sind unvermeidbar, sie müssen aber im Streben nach einem gemeinsamen Weg überwunden werden.} \\
\multicolumn{18}{l}{\textbf{Reference: } Conflict is inevitable, but must be overcome by the desire to walk together. } \\
 \textbf{Generation:}\quad

Step:& 1 & 2 & 3 & 4 & 5 & 6 & 7 & 8 & 9 & 10 & 11 & 12 & 13 & 14 & 15 & 16 & 17\\
Layer 1: \!\!\!&\!\!\! Conf-\!\!\! &\!\!\! lic-\!\!\! &\!\!\! ts \!\!\!&\!\!\! are        \!\!\!&\!\!\! inevitable \!\!\!&\!\!\! but \!\!\!&\!\!\! they \!\!\!&\!\!\! must \!\!\!&\!\!\! be   \!\!\!&\!\!\! overcome \!\!\!&\!\!\! the      \!\!\!&\!\!\! pursuit \!\!\!&\!\!\! for \!\!\!&\!\!\! a \!\!\!&\!\!\! common \!\!\!&\!\!\! path \!\!\!&\!\!\! . \\
Layer 2: \!\!\!&\!\!\! Conf-\!\!\!&\!\!\! lic-\!\!\!&\!\!\! ts \!\!\!&\!\!\! inevitable \!\!\!&\!\!\! ,          \!\!\!&\!\!\! but \!\!\!&\!\!\! they \!\!\!&\!\!\! must \!\!\!&\!\!\! be   \!\!\!&\!\!\! overcome \!\!\!&\!\!\! the      \!\!\!&\!\!\! pursuit \!\!\!&\!\!\! of  \!\!\!&\!\!\! a \!\!\!&\!\!\! common \!\!\!&\!\!\! path \!\!\!&\!\!\! . \\
Layer 3: \!\!\!&\!\!\! Conf-\!\!\!&\!\!\! lic-\!\!\!&\!\!\! ts \!\!\!&\!\!\! inevitable \!\!\!&\!\!\! inevitable \!\!\!&\!\!\! but \!\!\!&\!\!\! they \!\!\!&\!\!\! they \!\!\!&\!\!\! must \!\!\!&\!\!\! overcome \!\!\!&\!\!\! in       \!\!\!&\!\!\! pursuit \!\!\!&\!\!\! for \!\!\!&\!\!\! a \!\!\!&\!\!\! common \!\!\!&\!\!\! path \!\!\!&\!\!\! . \\
Layer 4: \!\!\!&\!\!\! Conf-\!\!\!&\!\!\! lic-\!\!\!&\!\!\! ts \!\!\!&\!\!\! inevitable \!\!\!&\!\!\! inevitable \!\!\!&\!\!\! but \!\!\!&\!\!\! but  \!\!\!&\!\!\! they \!\!\!&\!\!\! must \!\!\!&\!\!\! overcome \!\!\!&\!\!\! overcome \!\!\!&\!\!\! pursuit \!\!\!&\!\!\! for \!\!\!&\!\!\! a \!\!\!&\!\!\! common \!\!\!&\!\!\! path \!\!\!&\!\!\! . \\
Layer 5: \!\!\!&\!\!\! Conf-\!\!\!&\!\!\! lic-\!\!\!&\!\!\! ts \!\!\!&\!\!\! are        \!\!\!&\!\!\! inevitable \!\!\!&\!\!\! ,   \!\!\!&\!\!\! but  \!\!\!&\!\!\! they \!\!\!&\!\!\! must \!\!\!&\!\!\! be       \!\!\!&\!\!\! the      \!\!\!&\!\!\! pursuit \!\!\!&\!\!\! for \!\!\!&\!\!\! a \!\!\!&\!\!\! common \!\!\!&\!\!\! path \!\!\!&\!\!\! . \\
Layer 6: \!\!\!&\!\!\! Conf-\!\!\!&\!\!\! lic-\!\!\!&\!\!\! ts \!\!\!&\!\!\! are        \!\!\!&\!\!\! inevitable \!\!\!&\!\!\! ,   \!\!\!&\!\!\! but  \!\!\!&\!\!\! they \!\!\!&\!\!\! must \!\!\!&\!\!\! be       \!\!\!&\!\!\! the      \!\!\!&\!\!\! pursuit \!\!\!&\!\!\! for \!\!\!&\!\!\! a \!\!\!&\!\!\! common \!\!\!&\!\!\! path \!\!\!&\!\!\! .
\end{tabular}}\\
\hline
\end{tabular}}
\vspace{-.1cm}
\caption{Layer-wise predictions of our \model\ w/ vanilla NAT with examples in the test set of WMT'14 DE--EN translation. In our model, we use BPE segmentation, where prefixes are marked by - for the convenience of reading. A hyphen token is represented by ``@-@.'' }
\label{fig:case_study}
\vspace{-.2cm}
\end{figure*}

\textbf{Word Repetition.}
NAT models often generate repetitive tokens, because of the weakness in dealing with multi-modality \cite{gu2018nonautoregressive}, as expressions with slight different wordings can be thought of as a mode of the target distribution.
Following \citet{ghazvininejad-etal-2019-mask} and \citet{imputer}, 
we measure the fraction of repetitive tokens in each layer, shown in Table~\ref{fig:layer}b.

As seen, the repetition rate is high at early layers. However, it decreases drastically along the decoder layers. In general, the repetition rate is less than 15\% at the final layer for all three models. We also observe that a deeper model (12- or 18-layer) has fewer repetitions than a shallow model (6-layer). This further confirms that our layer-wise prediction is able to perform meaningful calibration and address the weakness of NAT models. 

\begin{figure}[!t]
\centering
\vspace{-.1cm}
\resizebox{\linewidth}{!}{
\mbox{\includegraphics[width=.5\linewidth]{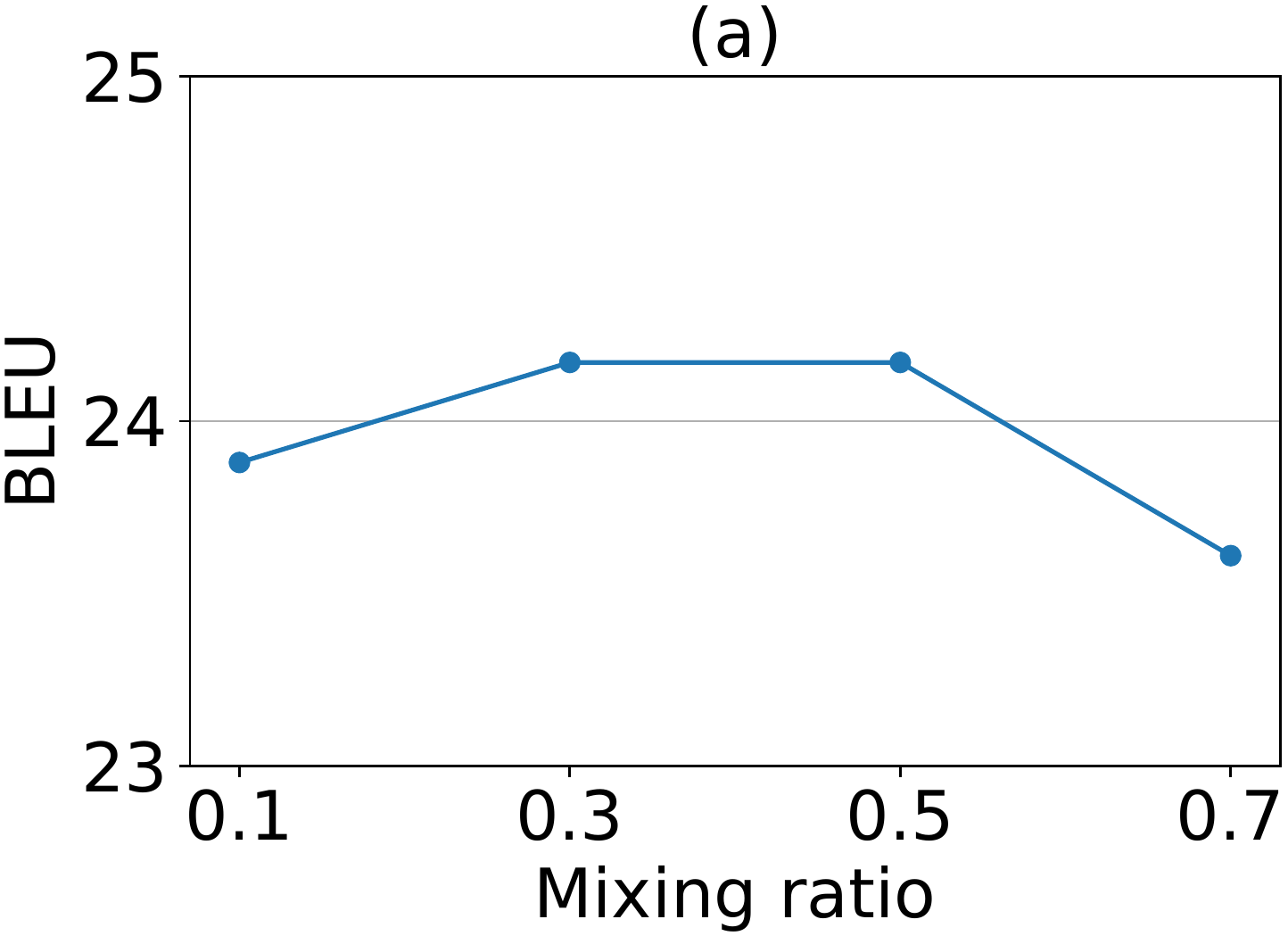}
\includegraphics[width=.52\linewidth]{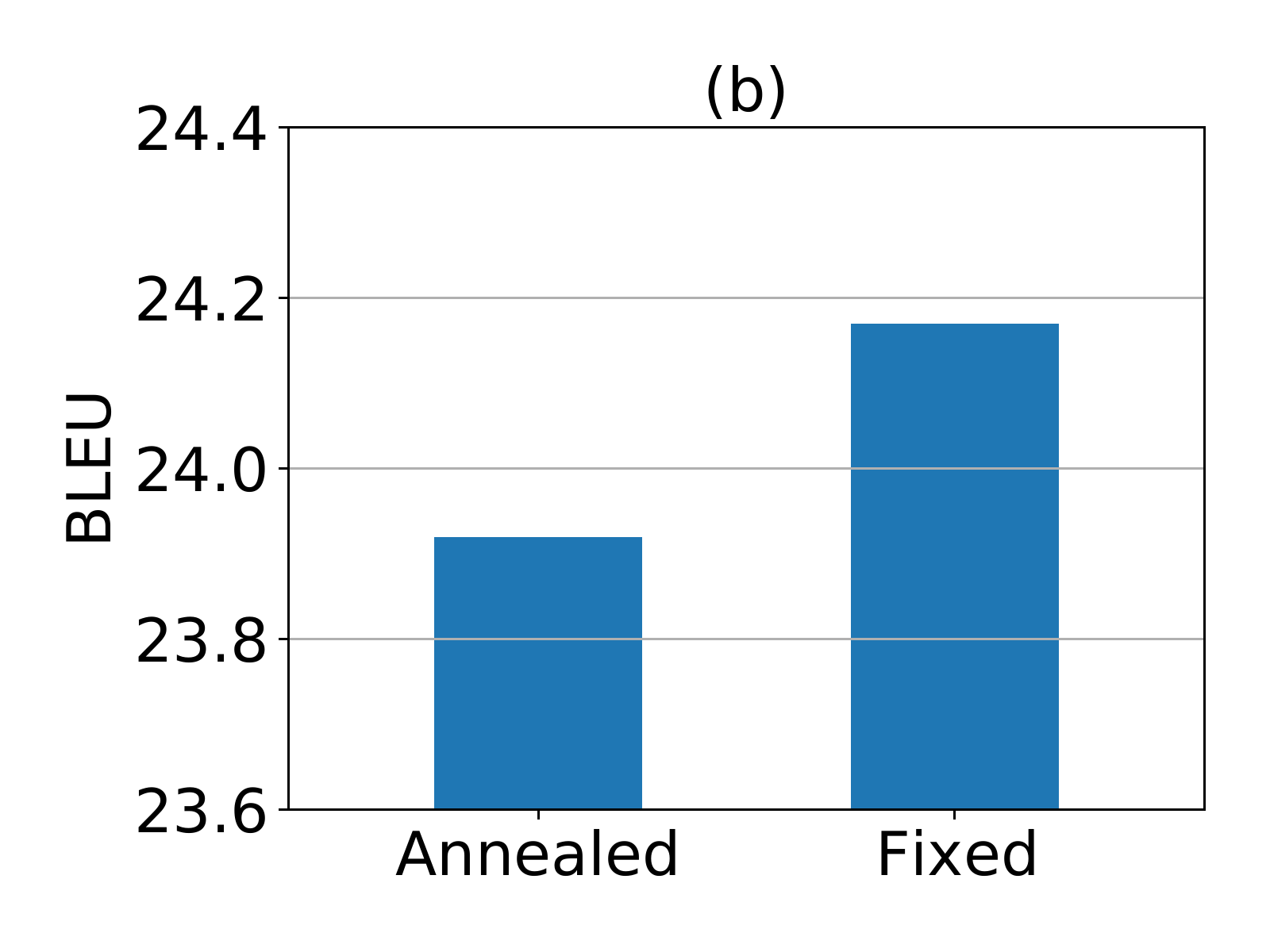}}}
 \caption{(a) Performance with different mixing ratios. (b) Impact of annealing the mixing ratio to zero during training.}
   \label{fig:layer_ss}
  \vspace{-0.1cm}
\end{figure}

\textbf{Mixed Training.}
 We tune the mixing ratio $\lambda$ and show the results in Figure~\ref{fig:layer_ss}a. We find that a mixing ratio around 0.3 performs well, but the performance degenerates significantly when the ratio is too high. This is because the intermediate predictions are important for the prediction awareness of follow-up layers; therefore the layer-wise predictions can not be fully replaced with groundtruth tokens. 
  
To better understand how the mixed training scheme works, we let the ratio of mixed groundtruth tokens anneal to zero throughout the training. In our experiments, we observe that with the annealed mixing ratio, the training eventually degenerates to that of the standard DSLP model, which results in worse final performance (see Figure~\ref{fig:layer_ss}b).
This phenomenon suggests that the intermediate predictions are not perfect during the entire training; consistently mixing the groundtruth provides correction to layer-wise predictions, which results in a better training. 
Based on our analysis, we set the mixing ratio as 0.3 during the entire training.

\subsection{Case Study}
\vspace{-.1cm}
In Figure~\ref{fig:case_study}, we demonstrate a case study on the test set of WMT'14 DE--EN dataset with our \model{}. Here, we present German-to-English translation for accessibility concerns.

As seen, the prediction by lower layers is of low quality, and calibration is indeed performed through the decoder layers. In Example~1, the words \textit{meet}, \textit{minded}, and \textit{minded} are repeated several times in Layer~1. This is expected, as there is no calibration in Layer~1. Then, our layer-wise prediction-aware decoder gradually revises out most of the repetitions. In Example~2, the phrase \textit{must be overcome the pursuit} is not fluent. The grammatical error is also corrected by the calibration process. 

Admittedly, the output of our \model\ may not be prefect. The last layer in Example~1 still has one repetition of the word \textit{dark-}, and in Example 2, the word \textit{overcome} is mistakenly dropped. We conjecture that this is because the target length is incorrectly predicted. This can probably be addressed with more decoder layers to find a paraphrase, or using other decoding methods such as noisy parallel decoding \cite[NPD,][]{gu2018nonautoregressive}. Like most translation models, our \model\ cannot perform non-literal translation well, and thus its output of Example 2 does not resemble the reference.

In general, our case study confirms that
the proposed \model{} model indeed performs calibration by layer-wise prediction-awareness, and that the translation quality is generally improved layer-by-layer.

\section{Conclusion}
In this paper, we propose a deeply supervised, layer-wise prediction-aware Transformer (\model{}) for NAT. Our approach is generic and can be combined with different NAT models. 
We conducted experiments on four machine translation tasks with four base models (vanilla NAT, CMLM, GLAT, and CTC). The results show that our method consistently improves the translation quality by more than 1 BLEU score, which is considered a large improvement over strong baselines. Our best variant achieves better BLEU scores than its autoregressive teacher model on three of four benchmark datasets, while accelerating the inference by 14.8x times. 

\bibliography{aaai22}

\begin{thebibliography}{26}
\providecommand{\natexlab}[1]{#1}

\bibitem[{Akaike(1969)}]{AR}
Akaike, H. 1969.
\newblock Fitting autoregressive models for prediction.
\newblock \emph{Annals of the Institute of Statistical Mathematics}, 21(1):
  243--247.

\bibitem[{Bahdanau, Cho, and Bengio(2015)}]{BahdanauCB14}
Bahdanau, D.; Cho, K.; and Bengio, Y. 2015.
\newblock Neural machine translation by jointly learning to align and
  translate.
\newblock In \emph{International Conference on Learning Representations}.

\bibitem[{Bao et~al.(2021)Bao, Huang, Xiao, Wang, Dai, and
  Chen}]{bao-etal-2021-non}
Bao, Y.; Huang, S.; Xiao, T.; Wang, D.; Dai, X.; and Chen, J. 2021.
\newblock Non-autoregressive translation by learning target categorical codes.
\newblock In \emph{Proceedings of the 2021 Conference of the North American
  Chapter of the Association for Computational Linguistics: Human Language
  Technologies}, 5749--5759.

\bibitem[{Chan et~al.(2020)Chan, Saharia, Hinton, Norouzi, and
  Jaitly}]{imputer}
Chan, W.; Saharia, C.; Hinton, G.; Norouzi, M.; and Jaitly, N. 2020.
\newblock Imputer: Sequence modelling via imputation and dynamic programming.
\newblock In \emph{Proceedings of the 37th International Conference on Machine
  Learning}, 1403--1413.

\bibitem[{Devlin et~al.(2019)Devlin, Chang, Lee, and
  Toutanova}]{devlin-etal-2019-bert}
Devlin, J.; Chang, M.-W.; Lee, K.; and Toutanova, K. 2019.
\newblock {BERT}: Pre-training of deep bidirectional transformers for language
  understanding.
\newblock In \emph{Proceedings of the 2019 Conference of the North {A}merican
  Chapter of the Association for Computational Linguistics: Human Language
  Technologies}, 4171--4186.

\bibitem[{Elbayad et~al.(2020)Elbayad, Gu, Grave, and Auli}]{depthadaptive}
Elbayad, M.; Gu, J.; Grave, E.; and Auli, M. 2020.
\newblock Depth-adaptive Transformer.
\newblock In \emph{International Conference on Learning Representations}.

\bibitem[{Ghazvininejad et~al.(2020)Ghazvininejad, Karpukhin, Zettlemoyer, and
  Levy}]{GhazvininejadKZ20}
Ghazvininejad, M.; Karpukhin, V.; Zettlemoyer, L.; and Levy, O. 2020.
\newblock Aligned cross entropy for non-autoregressive machine translation.
\newblock In \emph{Proceedings of the International Conference on Machine
  Learning}, 3515--3523.

\bibitem[{Ghazvininejad et~al.(2019)Ghazvininejad, Levy, Liu, and
  Zettlemoyer}]{ghazvininejad-etal-2019-mask}
Ghazvininejad, M.; Levy, O.; Liu, Y.; and Zettlemoyer, L. 2019.
\newblock Mask-predict: parallel decoding of conditional masked language
  models.
\newblock In \emph{Proceedings of the 2019 Conference on Empirical Methods in
  Natural Language Processing and the 9th International Joint Conference on
  Natural Language Processing}, 6112--6121.

\bibitem[{Graves et~al.(2006)Graves, Fern\'{a}ndez, Gomez, and
  Schmidhuber}]{ctc}
Graves, A.; Fern\'{a}ndez, S.; Gomez, F.; and Schmidhuber, J. 2006.
\newblock Connectionist temporal classification: labelling unsegmented sequence
  data with recurrent neural networks.
\newblock In \emph{Proceedings of the 23rd International Conference on Machine
  Learning}, 369--376.

\bibitem[{Gu et~al.(2018)Gu, Bradbury, Xiong, Li, and
  Socher}]{gu2018nonautoregressive}
Gu, J.; Bradbury, J.; Xiong, C.; Li, V.~O.; and Socher, R. 2018.
\newblock Non-autoregressive neural machine translation.
\newblock In \emph{International Conference on Learning Representations}.

\bibitem[{Gu and Kong(2021)}]{gu-kong-2021-fully}
Gu, J.; and Kong, X. 2021.
\newblock Fully non-autoregressive neural machine translation: tricks of the
  trade.
\newblock In \emph{Findings of the Association for Computational Linguistics:
  ACL-IJCNLP 2021}, 120--133.

\bibitem[{Gu, Wang, and Zhao(2019)}]{lev-transformer}
Gu, J.; Wang, C.; and Zhao, J. 2019.
\newblock Levenshtein Transformer.
\newblock In \emph{Advances in Neural Information Processing Systems}.

\bibitem[{Huang et~al.(2020)Huang, Perez, Ba, and Volkovs}]{pmlr-v119-huang20f}
Huang, X.~S.; Perez, F.; Ba, J.; and Volkovs, M. 2020.
\newblock Improving Transformer optimization through better initialization.
\newblock In \emph{Proceedings of the International Conference on Machine
  Learning}, 4475--4483.

\bibitem[{Kasai et~al.(2021)Kasai, Pappas, Peng, Cross, and
  Smith}]{kasai2021deep}
Kasai, J.; Pappas, N.; Peng, H.; Cross, J.; and Smith, N. 2021.
\newblock Deep encoder, shallow decoder: reevaluating non-autoregressive
  machine translation.
\newblock In \emph{International Conference on Learning Representations}.

\bibitem[{Lee, Mansimov, and Cho(2018)}]{lee-etal-2018-deterministic}
Lee, J.; Mansimov, E.; and Cho, K. 2018.
\newblock Deterministic non-autoregressive neural sequence modeling by
  iterative refinement.
\newblock In \emph{Proceedings of the 2018 Conference on Empirical Methods in
  Natural Language Processing}, 1173--1182.

\bibitem[{Ott et~al.(2019)Ott, Edunov, Baevski, Fan, Gross, Ng, Grangier, and
  Auli}]{fairseq}
Ott, M.; Edunov, S.; Baevski, A.; Fan, A.; Gross, S.; Ng, N.; Grangier, D.; and
  Auli, M. 2019.
\newblock {FairSeq}: a fast, extensible toolkit for sequence modeling.
\newblock In \emph{Proceedings of the 2019 Conference of the North {A}merican
  Chapter of the Association for Computational Linguistics (Demonstrations)},
  48--53.

\bibitem[{Papineni et~al.(2002)Papineni, Roukos, Ward, and
  Zhu}]{papineni-etal-2002-bleu}
Papineni, K.; Roukos, S.; Ward, T.; and Zhu, W.-J. 2002.
\newblock {BLEU}: A method for automatic evaluation of machine translation.
\newblock In \emph{Proceedings of the 40th Annual Meeting of the Association
  for Computational Linguistics}, 311--318.

\bibitem[{Qian et~al.(2021)Qian, Zhou, Bao, Wang, Qiu, Zhang, Yu, and
  Li}]{qian2020glancing}
Qian, L.; Zhou, H.; Bao, Y.; Wang, M.; Qiu, L.; Zhang, W.; Yu, Y.; and Li, L.
  2021.
\newblock Glancing Transformer for non-autoregressive neural machine
  translation.
\newblock In \emph{Proceedings of the 59th Annual Meeting of the Association
  for Computational Linguistics and the 11th International Joint Conference on
  Natural Language Processing}, 1993--2003.

\bibitem[{Saharia et~al.(2020)Saharia, Chan, Saxena, and
  Norouzi}]{saharia-etal-2020-non}
Saharia, C.; Chan, W.; Saxena, S.; and Norouzi, M. 2020.
\newblock Non-autoregressive machine translation with latent alignments.
\newblock In \emph{Proceedings of the 2020 Conference on Empirical Methods in
  Natural Language Processing}, 1098--1108.

\bibitem[{Stern et~al.(2019)Stern, Chan, Kiros, and Uszkoreit}]{stern}
Stern, M.; Chan, W.; Kiros, J.; and Uszkoreit, J. 2019.
\newblock Insertion Transformer: flexible sequence generation via insertion
  operations.
\newblock In \emph{Proceedings of the International Conference on Machine
  Learning}, 5976--5985.

\bibitem[{Stern, Shazeer, and Uszkoreit(2018)}]{blockwise}
Stern, M.; Shazeer, N.; and Uszkoreit, J. 2018.
\newblock Blockwise parallel decoding for deep autoregressive models.
\newblock In \emph{Advances in Neural Information Processing Systems},
  10107–10116.

\bibitem[{Sun et~al.(2019)Sun, Li, Wang, He, Lin, and Deng}]{SunLWHLD19}
Sun, Z.; Li, Z.; Wang, H.; He, D.; Lin, Z.; and Deng, Z. 2019.
\newblock Fast structured decoding for sequence models.
\newblock In \emph{Advances in Neural Information Processing Systems},
  3011--3020.

\bibitem[{Vaswani et~al.(2017)Vaswani, Shazeer, Parmar, Uszkoreit, Jones,
  Gomez, Kaiser, and Polosukhin}]{attentionisallyouneed}
Vaswani, A.; Shazeer, N.; Parmar, N.; Uszkoreit, J.; Jones, L.; Gomez, A.~N.;
  Kaiser, L.~u.; and Polosukhin, I. 2017.
\newblock Attention is all you Need.
\newblock In \emph{Advances in Neural Information Processing Systems}.

\bibitem[{Wang, Zhang, and Chen(2018)}]{wang-etal-2018-semi-autoregressive}
Wang, C.; Zhang, J.; and Chen, H. 2018.
\newblock Semi-autoregressive neural machine translation.
\newblock In \emph{Proceedings of the 2018 Conference on Empirical Methods in
  Natural Language Processing}, 479--488.

\bibitem[{Wei et~al.(2019)Wei, Wang, Zhou, Lin, and
  Sun}]{wei-etal-2019-imitation}
Wei, B.; Wang, M.; Zhou, H.; Lin, J.; and Sun, X. 2019.
\newblock Imitation learning for non-autoregressive neural machine translation.
\newblock In \emph{Proceedings of the 57th Annual Meeting of the Association
  for Computational Linguistics}, 1304--1312.

\bibitem[{Zhou, Gu, and Neubig(2020)}]{Zhou2020Understanding}
Zhou, C.; Gu, J.; and Neubig, G. 2020.
\newblock Understanding knowledge distillation in non-autoregressive machine
  translation.
\newblock In \emph{International Conference on Learning Representations}.

\end{thebibliography}

\end{document}